\documentclass{article}
\usepackage{ecltree}
\usepackage{epic}
\usepackage{eepic}
\usepackage{tree-dvips}

\title{Term Encoding of Typed Feature Structures}
\author{Dale Gerdemann
\\
\normalsize Seminar f\"ur Sprachwissenschaft, Universit\"at T\"ubingen
\\
\normalsize Kl. Wilhelmstr. 113
\\
\normalsize 72074 T\"ubingen, Germany
\\
\normalsize Email: dg@sfs.nphil.uni-tuebingen.de
}

\newcommand\type[1]{\mbox{\bf #1}}
\newcommand\feat[1]{\mbox{\sc#1}}
\newcommand\Sequence[1]{\langle#1\rangle}
\newcommand\PartialFunction[3]{#1\colon#2\rightharpoonup#3}

\newcommand\Type{\mbox{\sf Type}}
\newcommand\Sub{\sqsupseteq}
\newcommand\Feat{\mbox{\sf Feat}}
\newcommand\Approp{\mbox{\sf Approp}}

\newcommand\sref[1]{\S\ref{#1}}
\newcommand\fref[1]{fig.\ \ref{#1}}
\newcommand{\mcc}[2]{\multicolumn{#1}{c}{#2}}

\newcommand{\n}[2]{\node{#1}{#2}}
\newcommand{\nc}[2]{\nodeconnect[b]{#1}[t]{#2}}
\def\fs#1{$\left[\begin{tabular}{ll}#1\end{tabular}\right]$}
\def\fsx#1{\left[\begin{tabular}{ll}#1\end{tabular}\right]}

\begin{document}

\renewcommand{\baselinestretch}{1.15}

\maketitle
\thispagestyle{empty}

\section{Introduction}

I explore in this paper a variety of approaches to Prolog term
encoding typed feature structure grammars. As in Carpenter
\cite{carp:logi92}, the signature for such grammars consists of a
bounded complete partial order of types under subsumption ($\Sub$)
along with a partial function
$\PartialFunction{\Approp}{\Type\times\Feat}{\Type}$. The
appropriateness specification is subject to the constraint that
feature-value specifications for subtypes are at least as specific as
those for supertypes: if $Approp(\type{t},\feat{f}) = \type{s}$ and
\type{t} subsumes \type{t$'$}, then $Approp(\type{t$'$},\feat{f}) =
\type{s$'$}$ for some \type{s$'$} subsumed by \type{s}.\footnote{Other
  more complex constraints can be compiled out
  \cite{Goetz:Meurers:95}. So a system that only uses appropriateness
  conditions is more general than it might first appear.}

Previous approaches to term encoding of typed feature structures
(\cite{alsh:core91}, \cite{alsh:euro91}, \cite{erba:prof95}), have
assumed a similar signature plus additional restrictions such as:
limitations on multiple inheritance, or exclusion of more specific
feature-value declarations on subtypes. The encoding presented here is
subject to no such restrictions. The encoding will ensure that every
feature structure is well-typed (Carpenter \cite{carp:logi92}), i.e.,
for every feature \feat{f} on a node with type \type{t}, the value of this
feature must be subsumed by $Approp(\type{t},\feat{f})$. And furthermore, the
encoding will ensure, as required by {\sc hpsg}, that every feature
structure is extendible to a maximally specific well-type feature
structure.\footnote{See Pollard \& Sag \cite{poll:head94}, p.
  21\begin{quote} $\ldots$ a feature structure can be taken as a
  partial description of any of the well-typed (or totally well-typed,
  or totally well-typed and sort-resolved) feature structures that it
  subsumes. We choose to eliminate this possible source of confusion
  by using only totally well-typed sort-resolved feature structures as
  (total) models of linguistic entities and AVM diagrams (not feature
  structures) as descriptions.\end{quote} It follows from this, that
  for an AVM to describe something, it must be extendible to a totally
  well-typed sort-resolved (or {\em type-resolved}) feature structure.
  As is standard in computational linguistics, I use the term {\em
    feature structure} to mean what Pollard and Sag mean by AVM\@. It
  turns out that for computational purposes, we will never be
  interested in Pollard \& Sag's notion of a feature structure. For
  computational purposes, we want more compact feature structures,
  which can provably be extended to (totally) well-typed and
  type-resolved feature structures. See \cite{gerd:corr94}
  \cite{king:type94} for details.}

Previous approaches, discussed in \sref{type_as_path}, have adopted a
technique from Mellish \cite{mell:impl88} \cite{mell:term92} in which
each type is encoded as an open-ended data structure representing the
path taken through the type hierarchy to reach that type.  Or, in
other words, a type \type{t} is represented as a sequence of types,
starting at the most general type below \type{$\top$} and ending at
\type{t}, in which each consecutive pair consists of supertype
followed by immediate subtype. By bundling features together with the
types that introduce them, it is then possible to allow the number of
features on a type to increase as the type is further instantiated.
The disadvantage of this representation, though, is that there is no
unique path leading to a multiply-inherited type or any of its
subtypes. While a grammar with multiple-inheritance cannot generally
be represented in this approach, it is still possible, as explained in
\sref{compile_out_multi}, to compile the multiple inheritance out of
the type hierarchy and then term-encode a semantically equivalent
grammar. In this approach, multiple inheritance exists as a
convenience to the grammar writer, but is not actually used at run
time.

The encoding presented here will in some instances require the
introduction of disjunctions in order to ensure satisfiability of
feature structures, i.e., to ensure that feature structures are
extendible to maximally specific well-typed feature
structures.\footnote{I am simplifying here for the sake of exposition.
  The notion of ``satisfiable feature structure'' is treated in full
  in King \cite{king:type94}.} This introduction of disjunctions may,
in the worst case, exponentially increase the size of the grammar.
Practical experience with {\sc hpsg} grammars on the Troll system
(\cite{gerd:trol94}, \cite{hinr:part94}) has shown, however, that
feature structure compaction techniques can be used to keep this
increase reasonably small. In \sref{minimize}, I discuss how these
techniques can be incorporated into a term unification approach. In
\sref{unextend}, I discuss the technique of {\em unextension}, which
involves replacing disjunctions of feature structures with maximally
specific types on their nodes by smaller disjunctions of feature
structures in which the nodes are labelled by more general types. In
\sref{unfill}, I discuss the use of {\em unfilling} to remove
uninformative feature-value pairs. I show how this technique can be
used not only to make feature structures smaller, but also to
eliminate disjunctions. Then, in \sref{distribute}, I show how the
remaining disjunctions can sometimes be efficiently encoded as {\em
  distributed} (or {\em named}) disjunction. Distributed disjunctions
have been discussed elsewhere in the literature, but not in the
context of term encoded feature structures. Finally in
\sref{conclusions}, I summarize the approach taken in this paper and
discuss directions of future research.

\section{\label{type_as_path}Types-as-Paths Encoding}

\begin{figure}[htb]
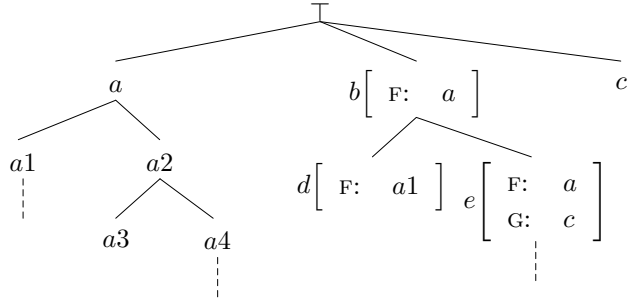
\centerline{
\begin{bundle}{$\top$}
  \GapWidth 30pt
  \chunk{\begin{bundle}{\raisebox{3pt}{$a$}}
      \chunk{\begin{bundle}{\raisebox{3pt}{$a1$}}
              \chunk{}
              \drawwith{\dashline[50]{3}}
              \end{bundle}}
      \chunk{\begin{bundle}{\raisebox{3pt}{$a2$}}
             \chunk{\raisebox{3pt}{$a3$}}
             \chunk{\begin{bundle}{\raisebox{3pt}{$a4$}}
              \chunk{}
              \drawwith{\dashline[50]{3}}
              \end{bundle}}
             \end{bundle}}
      \end{bundle}}
  \chunk{\begin{bundle}{$b$\fs{
                            \feat{f}:& $a$}}
     \chunk{$d$\fs{
                \feat{f}: & $a1$}}
     \chunk{\begin{bundle}{$e$\fs{
                \feat{f}: & $a$\\
                \feat{g}: & $c$}}
              \chunk{}
              \drawwith{\dashline[50]{3}}
             \end{bundle}}
     \end{bundle}}
   \chunk{\raisebox{3pt}{$c$}}
\end{bundle}}
\caption{\label{th1}A Simple Type Hierarchy and Appropriateness
  Specification}
\end{figure}

The types-as-paths encoding, introduced by Mellish \cite{mell:impl88},
uses an open-ended data structure representing the path taken through
the type hierarchy to reach that type. For simplicity, let us first
consider how to represent types that do not take any features, such
as the types subsumed by \type{a} in \fref{th1}.

\bigskip
\begin{tabular}{ll}
type\phantom{xxxx} & encoding\\ \hline
$a$ &{\tt a(\_)}\\
$a1$ &{\tt a(a1(\_))}\\
$a2$ &{\tt a(a2(\_))}\\
$a3$ &{\tt a(a2(a3))}\\
$a4$ &{\tt a(a2(a4(\_)))}
\end{tabular}
\bigskip

It is clear that the encodings for \type{a}, \type{a2} and \type{a4}
will all unify, whereas the encodings for \type{a2} and \type{a1} will
not unify.\footnote{The encoding used in {\sc ale} \cite{carp:ale94}
  is similar except that the paths may be gappy. Thus, \type{a4} could
  be encoded as any of the following: {\tt \_-a4-a2-a}, {\tt \_-a4-a},
  {\tt \_-a4-a2} or {\tt \_-a4}. In this example, the path reflects
  the derivational history of how the \type{a4} got to be an
  \type{a4}. This same principle is used in the implementation of
  updateable arrays in the Quintus Prolog library. Since the encoding
  is not unique, a special purpose unifier must be used, which
  dereferences each type before unifying. Thus this gappy
  representation is not applicable for the goal of this paper, which
  is to use ordinary (Prolog-style) term unification.} While, in
general, the encoding employs open data structures, the example shows
that \type{a3}, since it is maximally specific, can be encoded as a
ground term.  If, however, we wish to be able to distinguish in a
feature structure between reentrant and non-reentrant instances of
\type{a3}, then we would need encode \type{a3} also as an open term:
{\tt a(a2(a3(\_)))}.\footnote{\label{intensional}This idea is used,
  for example, in the ProFIT system \cite{erba:prof95} in order to
  distinguish between intensional and extensional types (see Carpenter
  \cite{carp:logi92} for this distinction). This distinction is
  certainly important.  However, as it is a side issue for this paper,
  I will ignore it.}

This encoding allows a particularly convenient encoding of feature
information. If a type introduces $n$ features, then we simply add $n$
additional argument slots to that type in each path. For example,
consider the encodings of \type{b} and \type{e}:
\bigskip

\begin{tabular}{ll}
type\phantom{xxxx} & encoding\\ \hline
$b$ &{\tt b(\_,\_F))}\\
$e$ &{\tt b(e(\_,\_G),\_F))}
\end{tabular}
\bigskip

As can be seen from this example, there is essentially no difference
between the encoding of a simple type \type{t} and the encoding of a
feature structure of type \type{t}. The encoding of each type has slots in
it where all of the feature value information can be included. The
encoding for \type{e} is particularly instructive. As can be seen, \type{e}
takes two features, which are introduced at two different points along
the path. Furthermore, \type{e} is a subtype of \type{b}, which has only a
single argument position for the feature \feat{f}. It is clear then that
the number of slots for features can increase as a type is further
instantiated.

The idea of bundling feature information along with the type that
introduces that feature is certainly elegant. However, there is a
drawback when not all of the information about a feature is located on
a single type. For example, the type \type{d} inherits a feature
\feat{f} from the supertype \type{b}, but then adds a more specific
value specification for this feature. Consider what happens when the
feature structure $b[\feat{f}:~a]$ (encoded: {\tt b(\_,a(\_)))})
unifies with the feature structure $d$ (encoded: {\tt b(d,\_F))}). The
unification of these two encodings is {\tt b(d,a(\_)))}, i.e.,
$d[\feat{f}:~a]$, which is not well-typed. This, however, is not
really a problem specific to the types-as-paths encoding. Any typed
feature structure approach with appropriateness specifications needs
to incorporate type inferencing.  As discussed in \sref{closed_world}
(see also \cite{gerd:corr94}), one way to maintain appropriateness is
to multiply out disjunctive possibilities. For example,
$b[\feat{f}:~a]$ should be multiplied out to: $\{d[\feat{f}:~a1],
e[\feat{f}:~a]\}$. As seen in the next section, handling multiple
inheritance will also involve introducing disjunctions. So the problem
of efficiently representing such disjunction (discussed in
\sref{minimize}) will be of crucial importance.

\subsection{\label{compile_out_multi}Compiling Out Multiple Inheritance}

Multiple inheritance is a genuine problem for the types-as-paths
representation. The problem is that if a type can be reached by
multiple paths through the type hierarchy, then there is no longer a
unique representation for that type. A partial solution to this to use
{\em multi-dimensional inheritance} as in Erbach \cite{erba:prof95}
and Mellish \cite{mell:impl88}. This idea involves encoding types as a
set of paths rather than as a single path. The intuition is supposed
to be that each path in this set represents a different dimension in
the inheritance hierarchy.

\begin{figure}[htb]
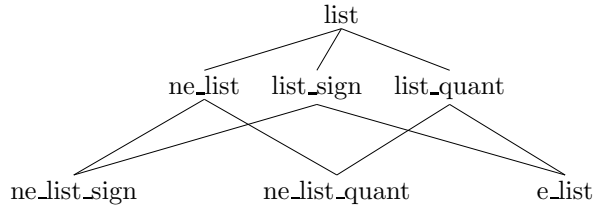
\centerline{
\begin{tabular}[t]{ccccccc}
\mcc{7}{\n{list}{list}}\\[3ex]
& \mcc{2}{\n{nelist}{ne\_list}} & \n{listsign}{list\_sign} &
  \mcc{2}{\n{listquant}{list\_quant}}\\[6ex]
\n{nelistsign}{ne\_list\_sign} &
      \mcc{5}{\n{nelistquant}{ne\_list\_quant}} & \n{elist}{e\_list}
\nc{list}{nelist} \nc{list}{listsign} \nc{list}{listquant}
\nc{listsign}{elist} \nc{listquant}{elist} \nc{nelist}{nelistquant}
\nc{listquant}{nelistquant} \nc{nelist}{nelistsign}
\nc{listsign}{nelistsign}
\end{tabular}}
\caption{\label{listth}Multiple inheritance in type hierarchy for lists}
\end{figure}

Consider, for example, the commonly used type hierarchy for lists in
\fref{listth}.
The types \type{ne\_list}, \type{list\_sign} and \type{list\_quant}
are not mutually exclusive, so it seems reasonable to represent these
types by a set of paths. \type{ne\_list}, for example, could be
represented as {\tt list(ne\_list,\_,\_)} and \type{ne\_list\_quant}
as {\tt list(ne\_list,\_,list\_quant)}. This seems to work fine except
that there would be no way to rule out the unification of the
encodings for \type{ne\_list\_quant} and \type{ne\_list\_sign}. These
two types unify to give the non-existent type {\tt
  list(ne\_list,list\_sign,list\_quant)}. So the restriction on this
encoding is the following: if types \type{t} and \type{t$'$}
have a subtype in common, then every subtype of \type{t} must
be subsumed by \type{t$'$}. In the \type{list} example,
\type{ne\_list} and \type{list\_sign} have a subtype in common
(\type{ne\_list\_sign}). However, \type{list\_sign} also subsumes
\type{e\_list}, which is not subsumed by \type{ne\_list}. So while
this approach may work for some special cases of multiple inheritance,
it is not a general solution to the problem.\footnote{A second problem
  with the approach is that it provides no way to attach features to
  multiply inherited types. So if \type{ne\_list\_quant} has some
  features which are not inherited from either \type{ne\_list} or
  \type{list\_quant}, then there is no position in either path of
  types to which these features can be attached.}

\begin{figure}[htb]
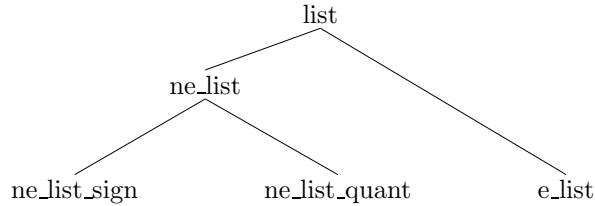
\centerline{
\begin{tabular}[t]{ccccccc}
\mcc{7}{\n{list}{list}}\\[3ex]
& \mcc{2}{\n{nelist}{ne\_list}} & \n{listsign}{\phantom{list\_sign}} &
   \mcc{2}{\n{listquant}{\phantom{list\_quant}}}\\[6ex]
\n{nelistsign}{ne\_list\_sign} &
      \mcc{5}{\n{nelistquant}{ne\_list\_quant}} & \n{elist}{e\_list}
\nc{list}{nelist}
\nc{list}{elist} \nc{nelist}{nelistquant}
\nc{nelist}{nelistsign}
\end{tabular}}
\caption{\label{th2}Type hierarchy for lists with multiple
  inheritance compiled out}
\end{figure}

Given the problems with representing multiple inheritance, it is
reasonable to ask how important multiple inheritance is. In fact,
given some reasonable closed world assumptions as discussed in
the next section, it is possible to compile out all multiple
inheritance. To see this, consider what happens when we remove from
the list hierarchy the types \type{list\_sign} and \type{list\_quant}. As seen
in \fref{th2},
removing these two types is sufficient to eliminate all multiple
inheritance. In order to legitimize removing types from the type
hierarchy, two further steps are needed. First, some disjunctive
appropriateness specifications must be introduced. For example, for
any type \type{t} and feature \feat{f} with $Approp(\type{t},\feat{f})
= \type{list\_sign}$, the new appropriateness conditions should
include the disjunctive specification $Approp(\type{t},\feat{f}) =
\{\type{ne\_list\_sign},
\type{e\_list}\}$.\footnote{Such disjunctive appropriateness specifications
  are allowed, at least internally, in the Troll system
  \cite{gerd:trol94}.} And second, any feature structure containing
the type \type{list\_sign} on a node must be compiled into two feature
structures: one with \type{ne\_list\_sign} and one with
\type{e\_list}. So some lexical entries or rules may have to be
compiled out into multiple instances. The question of how to deal with
this introduced disjunction is treated in \sref{minimize}.

\subsection{\label{closed_world}Closed World Assumptions}

So far, we have seen that a closed world interpretation of the type
hierarchy will allow us to compile out multiple inheritance at the
cost of introducing some disjunctions into the grammar. But the closed
world assumption is not a condition that is imposed solely for the
purpose of term encoding. As shown in Gerdemann \& King
\cite{gerd:type93} \cite{gerd:corr94}, the closed world assumption is
needed if types are to be used to encode any kind of feature
cooccurrence restrictions. As noted by Copestake et al.\
\cite{cope:acqu93} this deficiency of open-world type systems leads to
serious problems for expressing any kind of linguistic constraints.

\begin{figure}[htb]
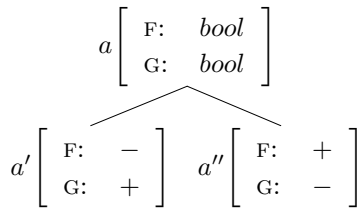
\centerline{
\begin{bundle}{$a$\fs{
                \feat{f}:& $bool$\\
                \feat{g}:& $bool$}}
  \GapWidth 10pt
     \chunk{$a'$\fs{
                \feat{f}: & $-$\\
                \feat{g}: & $+$}}
     \chunk{$a''$\fs{
                \feat{f}: & $+$\\
                \feat{g}: & $-$}}
     \end{bundle}}
\caption{\label{th3}Type hierarchy requiring type resolution}
\end{figure}

While the closed world assumption is clearly needed, it is also the
case that there is a price to be paid for maintaining this condition.
In particular, Gerdemann \& King \cite{gerd:type93} \cite{gerd:corr94}
showed that maintaining this condition will sometimes involve
multiplying out disjunctive possibilities. For example, consider the
type hierarchy in \fref{th3}.
Every feature structure of type \type{a} must ultimately be resolved
to either a feature structure of type \type{a$'$} or of type
\type{a$''$}.  So the feature structure $a[\feat{f}\kern -2pt :~bool,
\feat{g}\kern -2pt :~bool]$ really represents the set of resolvants
$\{a'[\feat{f}\kern -2pt :~+, \feat{g}\kern -2pt :~-],
a''[\feat{f}\kern -2pt :~-, \feat{g}\kern -2pt :~+]\}$ and the feature
structure $a[\feat{f}\kern -2pt :~\framebox{\scriptsize 1},
\feat{g}\kern -2pt :~\framebox{\scriptsize 1}\kern 3pt]$ really
represents the empty set of feature structures.\footnote{Note that
  this last feature structure would be considered ``well-typed'' in
  {\sc ale}. For {\sc ale}, one must assume an open world semantics in
  which an object described by $a[\feat{f}\kern -2pt
  :~\framebox{\scriptsize 1}, \feat{g}\kern -2pt
  :~\framebox{\scriptsize 1}\kern 3pt]$ is neither of type \type{a$'$}
  nor of type \type{a$''$}.} To make our terminology precise, let us
call feature structures such as $a'[\feat{f}\kern -2pt :~+],
a'[\feat{f}\kern -2pt :~-], a''[\feat{f}\kern -2pt :~+]$ and
$a''[\feat{f}\kern -2pt :~-]$ {\em extensions} of the feature
structure $a[\feat{f}\kern -2pt :~bool]$. So a feature structure
$\alpha$ can be extended by replacing the type on each node by a
species subsumed by that type. The set of resolvants is then the set
of extensions which also satisfy the appropriateness conditions: in
this case $\{a'[\feat{f}\kern -2pt :~+], a''[\feat{f}\kern -2pt
:~-]\}$. Note that the resolvants of a feature structure need not be
totally well typed in the sense of Carptenter \cite{carp:logi92},
i.e., there can be appropriate features such as \feat{g} which are not
in the resolvant.\footnote{In fact, in general resolved feature
  structures cannot be totally well typed since total-well typing
  creates more nodes in a feature structure which would then need to
  be resolved and then total-well typed again. The result is an
  infinite loop \cite{gerd:corr94}.} This fact will be of great
importance when we consider efficient representations for sets of
resolvants in the next section.

So if our term representation for typed feature structures is to
maintain the constraints imposed by the appropriateness specification,
it looks as if we will again have to expand feature structures out
into multiple instances. In fact, as discussed in \sref{minimize},
there are methods both for reducing the need for disjunctions and for
compactly representing those disjunctions that can't be eliminated.
But before considering these methods, let us first consider one
very desirable property of resolved feature structures, namely that
the resolved feature structures are closed under unification.

If we use a specialized feature structure unifier, then there exists
the possibility of building in type inferencing as part of
unification. If all unification is simply term unification, however,
then we don't have this option. We need all type inferencing to be
static, i.e., applied at compile time.\footnote{Note, for the sake of
  comparison, that type inferencing is not static in {\sc ale}. The
  unification of well-typed feature structures is not guaranteed to be
  well-typed. It is, in fact, not even guaranteed to be well-typable.
Thus, the {\sc ale} unifier must do type inferencing at run time.}
So when two feature structures which satisfy appropriateness
conditions unify, the result of this unification must also satisfy
appropriateness conditions. As shown in Gerdemann \& King
\cite{gerd:corr94} and King \cite{king:type94}, this is indeed the
case for resolved feature structures. Intuitively, the reason for this
is quite simple. Type inferencing in a system like {\sc ale} has the
effect of increasing the specificity of types on certain nodes in a
feature structure. If all of these types are already maximally
specific, then {\sc ale}-style type inferencing could only apply
vacuously.

\section{\label{minimize}Minimizing Disjunctions}

We have now seen two instances in which feature structures may need to
be multiplied out into disjunctive possibilities. First, this may
arise as a result of eliminating multiple inheritance from the type
hierarchy. And second, as a result of the type resolution which is
needed in order to ensure static typing. Now in this section, I
discuss how the need for this disjunction can be reduced by the
technique of {\em unextension} \sref{unextend} and by {\em unfilling}
\sref{unfill}. And then in \sref{distribute}, I show how the remaining
disjunctions can at least sometimes be efficiently represented by
using distributed disjunctions.

\subsection{\label{unextend}Unextension}

First, consider the disjunction that is introduced by type resolution.
Most of this disjunction can be eliminated by using the technique of
{\em unextension}. Recall that in \sref{closed_world}, I defined the
extension of a feature structure $\alpha$ to be a feature structure in
which the type on each node of $\alpha$ has been replaced by a species
subsumed by that type. Now, let us define the extensions of a set of
feature structures $S$ as the set of all extensions of the members of
$S$. And then define the unextension of a set of feature structures
$S$ as the minimal cardinality set of feature structures $S'$ such
that $extensions(S') = S$.\footnote{\label{fn}The normal form result
  of G\"otz \cite{gotz:norm93} shows, albeit rather indirectly, that
  there is a unique unextension for the set of resolvants of a feature
  structure---assuming there are no unary branches in the type
  hierarchy.  For G\"otz, unextension is an implicit part of his
  function for resolving feature structures. I have abstracted
  unextension out as a separate operation purely for expository
  reasons. In an actual implementation (such as Troll
  \cite{gerd:trol94}) it makes more sense to follow G\"{o}tz's
  approach since it is not very efficient for the compiler to expand
  feature structures out into huge sets that then have to be collapsed
  back down.} So whenever a feature structure is resolved to a large
disjunction of feature structures $S$, we can normally compact this
disjunction back down to a much smaller unextended set $S'$.  Since
the extensions of $S$ and $S'$ are the same, it is clear that these
two sets of feature structures represent the same information.

As a simple---but extremely commonly occurring---example, consider a
feature structure $F$, all of whose extensions are well-typed. In this
case, the set of resolvants of $F$ is exactly the set of extensions of
$F$. So the unextended set of resolvants of $F$ is simply
$\{F\}$.\footnote{There is, however, one complication, namely, if
  there are unary branches in the type hierarchy, then a feature
  structure might be resolved and then unextended back to a slightly
  different, but semantically identical feature structure (see
  Carpenter \cite{carp:logi92}, chap. 9). For our purposes here, it
  doesn't matter if the unextension of a set of feature structures is
  not unique.} So resolving and then
unextending $F$ appears at first to accomplish nothing. But, in fact,
there is a gain. Initially, with the feature structure $F$ we had no
guarantee that static typing would be safe.  But with the
resolved-unextended set $\{F\}$, we now know that there are no non-well
typed extensions, so there will never be a need for run-time type
inferencing.

As another example, consider the set of resolvants that {\sc hpsg}
allows for:
\begin{quote}
$head$-$struc[\feat{head-dtr}\kern -2pt : sign,\:%
 \feat{comp-dtrs}\kern -2pt : elist]$
\end{quote}
Since the only subtype of \type{head-struc} for which \type{elist} is
an appropriate value on \type{comp-dtrs} is \type{head-comp-struc}, we
get the following set of resolvants:
\begin{quote}
$\{head$-$comp$-$struc[\feat{head-dtr}\kern -2pt : word,\:%
\feat{comp-dtrs}\kern -2pt : elist],\\
   \phantom{\{ }head$-$comp$-$struc[\feat{head-dtr}\kern -2pt : phrase,
\: \feat{comp-dtrs}\kern -2pt : elist]\}$.
\end{quote}
This two element set corresponds exactly to the set of extensions for
the following one element unextended set:
\begin{quote}
$\{head$-$comp$-$struc[\feat{head-dtr}\kern -2pt : sign,\:
 \feat{comp-dtrs}\kern -2pt : elist]\}$
\end{quote}
So the combination of resolving and unextending simply has the effect
of bumping the top-level type from \type{head-struc} to
\type{head-comp-struc}. This then rules out the malformed feature
structure that might have been obtained, for example, by unification with a
feature structure of type \type{head-filler-struc}.

Consider now the disjunctions that are introduced by compiling out
multiple inheritance. Recall that this technique involves eliminating
intermediate types from the hierarchy. It does not remove any species
from the hierarchy. Thus, regardless of whether or not multiple
inheritance has been compiled out, the resolvants of a feature
structure $F$ will be exactly the same. Since, unextension involves
replacing species on nodes with intermediate types, the possibilities
for unextending a set of feature structures will be reduced when some
intermediate types have been removed. So it turns out that compiling
out multiple inheritance actually introduces disjunction in a rather
indirect manner.

\subsection{\label{unfill}Unfilling}

Another operation that can be used to reduce the need for disjunction
is {\em unfilling}, which is the reverse of Carpenter's
\cite{carp:logi92} fill operation. In general, the purpose of
unfilling is to keep feature structures small. If a feature in a
feature structure has a value which is no more specific than the
appropriateness specification would require, then---assuming no
reentrancies would be eliminated and no dangling parts of the feature
structure would be created---that feature and its value may be
removed. So, in {\sc hpsg} for example, if the \feat{aux} feature in a
feature structure of type \type{verb} has the value \type{boolean}, then this
\feat{aux} feature can be removed.

Unfilling is most important, however, when it turns out that
eliminating features allows us to further apply unextension to
eliminate disjunctions. For example, given the type hierarchy and
appropriateness specification in \fref{th3} above, the feature
structure $a[\feat{f}:~bool, \feat{g}:~bool]$ is resolved to
$\{a'[\feat{f}:~+, \feat{g}:~-], a''[\feat{f}:~-, \feat{g}:~+]\}$.
However, both of the features \feat{f} and \feat{g} have uninformative
values, i.e., values which tell us nothing more than we already know
from the appropriateness specification. In fact, this is true both in
the unresolved feature structure and in each of the resolvants. So
these two features can be unfilled to give us the new set of
resolvants $\{a', a''\}$. This new set of resolvants can now be
unextended to the set $\{a\}$.

There is clearly a great deal more that can be said about unfilling
and about the class of unfilled feature structures. The issues,
however, are not specific to the problem of term encoding. So I
will simply refer the reader to the discussion in Gerdemann \& King
\cite{gerd:corr94} and G\"otz \cite{gotz:norm93}.

\subsection{\label{distribute}Distributed Disjunctions}

Unextension and unfilling can be used to eliminate quite a lot of
disjunctions. Unfortunately, not all disjunctions can be eliminated
with these techniques. But still, as noted in Gerdemann \& King
\cite{gerd:type93} \cite{gerd:corr94}, the remaining disjunctions,
have a rather special property. All of the resolvants of a feature
structure have the same shape. They differ only in the types labelling
the nodes. In a graph-unification based approach, this property can be
used to allow the use of a relatively simple version of {\em
  distributed} (or {\em named}) disjunction. This device can be used
to push top level disjunctions down to local, interacting disjunctions
as in this example:\footnote{\label{unrealistic}The example is a
  little unrealistic since the features \feat{head} and \feat{tail}
  could be unfilled. To make the example more realistic, imagine the
  feature structures embedded in a larger feature structure and
  imagine that the values of \feat{head} and \feat{tail} are reentrant
  with other parts of this feature structure, so that these features
  could not be unfilled without breaking these reentrancies.}

\begin{center}
$
\left\{
\fsx{$ne\_list\_sign$\\
    \feat{head}: $sign$\\
    \feat{tail}:  $list\_sign$\kern -2pt},
\fsx{$ne\_list\_quant$\\
    \feat{head}:  $quant$\\
    \feat{tail}:  $list\_quant$\kern -2pt}
\right\}
\Rightarrow
\fsx{$\Sequence{1\:ne\_list\_sign\:ne\_list\_quant}$\\
    \feat{head}:  $\Sequence{1\:sign\:quant}$\\
    \feat{tail}:  $\Sequence{1\:list\_sign\:list\_quant}$}
$
\end{center}

\noindent The idea is that if the $n$th alternative is chosen for a
disjunction of a particular name, then then $n$th alternative has to
be chosen uniformly for all other instances of disjunction with the
same name.

Distributed disjunctions have been discussed in a fair amount of
recent literature (\cite{maxw:over89}, \cite{dorr:feat90},
\cite{gerd:pars91}, \cite{grif:opti95}). This version of distributed
disjunction, however, is particularly simple since only the type
labels are involved. It is not at all difficult to modify a graph
unification algorithm in order to handle such disjunctions. Such a
modified unification algorithm is used, for example, in the Troll
system \cite{gerd:trol94}.

For term-represented feature structures, however, it will not be
possible to directly encode such distributed
disjunctions.\footnote{Actually there is an alternative representation
  for types that would allow a limited amount of direct encoding of
  distributed disjunctions. Following the basic idea of Mellish
  \cite{mell:impl88}, one could represent each type as a set of
  species encoded as a $vset$, where a vset is a term
  $vset(X_0,X_1,\ldots,X_N)$, with the conditions that:
\begin{itemize}
\item $X_0$ = 0
\item $X_N$ = 1
\item $X_i$ = $X_{i-1}$ if the $i$'th possible element is not in the set
\end{itemize}
With this {\it types-as-vsets} representation, some dependencies can
be represented as variable sharing across vsets. This idea is
elaborated upon in an earlier (and longer) version of the present
paper.} So rather than encoding distributed disjunctions in a feature
structure, we must encode them as definite clause attachments to the
feature structure. The idea is fairly simple, to efficiently represent
a set of feature structures $S$, we factor $S$ into first
term-represented feature structure, $\sqcup S$, expressing the
commonalities across all of the members of $S$, and second, a set of
definite clause attachments expressing all of the allowable further
extensions. There is, however, one hitch; namely, how do we know that
$\sqcup S$ will be expressible as a Prolog term? If $\sqcup S$ is well
typed, then there is no problem. But if $\sqcup S$ contains a node
labeled with \type{t} with a with an inappropriate feature \feat{f},
then the types-as-paths representation simply provides no argument
position for this inappropriate feature.

A solution to this problem can be found by imposing the feature
introduction condition of Carpenter \cite{carp:logi92}. This condition
requires that for each feature \feat{f}, there is a unique most
general type $Intro(\feat{f})$ for which this feature is appropriate.
Or, the other way around, if \feat{f} is appropriate for \type{t} and
\type{t$'$}, then it will also be appropriate for $\type{t} \sqcup
\type{t$''$}$. It is straightforward to see, then, that given the
feature introduction condition, if the feature structures $FS$ and
$FS'$ contain no inappropriate features, then $FS \sqcup FS'$ will
also contain no inappropriate features.

As an example, consider again the type hierarchy in \fref{th1}.
Suppose that we want to efficiently represent the set:\footnote{Again,
  this is an unrealistic example (see footnote \ref{unrealistic}).
  One would not normally need distributed disjunctions for such a simple
  case.}
\begin{quote}
$\{d[\feat{f}\kern -2pt : a1], e[\feat{f}\kern -2pt : a]\}$
\end{quote}
The generalization of these two feature structures is then term
encodable as follows:
\begin{quote}
$b[\feat{f}\kern -2pt : a]$ $\equiv$ {\tt b(X, a(Y))}
\end{quote}
Suppose now, that we want to use this feature structure as the lexical
entry for the word $w$. For this simple example, we can encode this
with just one definite clause attachment:\footnote{Multiple
  attachments would correspond to named disjunctions with different names.}
\begin{quote}
\begin{verbatim}
lex(w, b(X, a(Y))) :-
   p(X, Y).

p(d, a1(_)).
p(e(_), _).
\end{verbatim}
\end{quote}
One should note here the similarity to distributed disjunctions. The
term {\tt b(X, a(Y))} represents the underlying shape of the feature
structure and the defining clauses for {\tt p} encode dependencies
between types. It is, in fact, rather surprising that this division is
even possible. One of the features of the types-as-paths encoding is
that the features are bundled together with the types that introduce
them. So one might not have expected to see these two types of
information unbundled in this manner.

\section{\label{conclusions}Conclusions}

Some previous approaches to term encoding of typed feature structures
have enforced restrictions against multiple inheritance and against
having having more specific feature-value declarations on subtypes.
But such restrictions make typing virtually useless for encoding any
meaningful constraints. The only restriction imposed in the present
approach is the feature introduction condition, which is also
imposed in {\sc ale} (\cite{carp:ale94}). In fact, even this minor
restriction could be eliminated if we were to allow somewhat less
efficient definite clause attachments.

We have seen that in order to enforce constraints encoded in the
appropriateness specifications, it will sometimes be necessary to use
definite clause attachments to encode disjunctive possibilities.
Practical experience with the Troll system suggests that not many such
attachments will be needed. Nevertheless, they will arise and will
therefore need to be processed efficiently. Certainly options such as
delaying these goals or otherwise treating them as constraints can be
explored. In fact, one of the main advantages of having a term
encoding is that so many options are available from all of the
literature on efficient processing of logic programs. So the approach
to term encoding presented here should really be viewed as just the
first step in the direction of efficient processing of typed feature
structure grammars.

\end{document}